\documentclass[review,12pt,authoryear]{elsarticle}

\makeatletter
\def\ps@pprintTitle{%
    \let\@oddhead\@empty
    \let\@evenhead\@empty
    \def\@oddfoot{}
    \let\@evenfoot\@oddfoot
    }
\makeatother
\usepackage{apalike}
\usepackage[utf8]{inputenc}
\usepackage{multirow}
\usepackage{booktabs} 
\usepackage{times}  
\usepackage{helvet}  
\usepackage{courier}  
\usepackage{url}  
\usepackage{graphicx}  
\usepackage{amsfonts}
\usepackage{comment}
\usepackage{algorithm}
\usepackage{algorithmic}
\usepackage{amsmath}
\usepackage{cleveref}
\usepackage{subcaption}
\usepackage{epstopdf}

\usepackage{color}

\newcommand{\firstsite}{\textit{Koumbia}}
\newcommand{\secondsite}{\textit{Centre Val de Loire}}
\newcommand{\method}{\textit{REFeD}}

\begin{document}

\begin{frontmatter}

\title{Reuse out-of-year data to enhance land cover mapping via feature disentanglement and contrastive learning}
%
%
%

\author[inrae,inria]{C\'{a}ssio F. Dantas\corref{mycorrespondingauthor}}
\ead{cassio.fraga-dantas@inrae.fr}
\author[cirad,inria]{Raffaele Gaetano}
\ead{raffaele.gaetano@cirad.fr}
\author[twente]{Claudia Paris}
\ead{c.paris@utwente.nl}
\author[inrae,inria]{Dino Ienco}
\ead{dino.ienco@inrae.fr}
\address[inrae]{INRAE, UMR TETIS, Univ. Montpellier, Montpellier, France}
\address[cirad]{CIRAD, UMR TETIS, Univ. Montpellier, Montpellier, France}
\address[inria]{INRIA, Montpellier, France}
\address[twente]{Department of Natural Resources, ITC, University of Twente, Enschede, Netherlands}

\cortext[mycorrespondingauthor]{Corresponding author}

\begin{abstract}
Timely up-to-date land use/land cover (LULC) maps play a pivotal role in supporting agricultural territory management, environmental monitoring and facilitating well-informed and sustainable decision-making. Typically, when creating a land cover (LC) map, precise ground truth data is collected through time-consuming and expensive field campaigns. This data is then utilized in conjunction with satellite image time series (SITS) through advanced machine learning algorithms to get the final map. Unfortunately, each time this process is repeated (e.g., annually over a region to estimate agricultural production or potential biodiversity loss), new ground truth data must be collected, leading to the complete disregard of previously gathered reference data despite the substantial financial and time investment they have required.
How to make value of historical data, from the same or similar study sites, to enhance the current LULC mapping process constitutes a significant challenge that could enable the financial and human-resource efforts invested in previous data campaigns to be valued again. 
Aiming to tackle this important challenge, we here propose a deep learning framework based on recent advances in domain adaptation and generalization to combine remote sensing and reference data coming from two different domains (e.g. historical data and fresh ones) to ameliorate the current LC mapping process. 
Our approach, namely \method{} (data Reuse with Effective Feature Disentanglement
for land cover mapping), leverages a disentanglement strategy, based on contrastive learning, where invariant and specific per-domain features are derived to recover the intrinsic information related to the downstream LC mapping task and alleviate possible distribution shifts between domains. Additionally, \method{} is equipped with an effective supervision scheme where feature disentanglement is further enforced via multiple levels of supervision at different granularities.
The experimental assessment over two study areas covering extremely diverse and contrasted landscapes, namely \firstsite{} (located in the West-Africa region, in Burkina Faso) and \secondsite{} (located in centre Europe, France), underlines the quality of our framework and the obtained findings demonstrate that out-of-year information coming from the same (or similar) study site, at different periods of time, can constitute a valuable additional source of information to enhance the LC mapping process.
\end{abstract}

\end{frontmatter}

\section{Introduction}
The unprecedented availability of Earth observation (EO) information regularly acquired through modern public and private EO Programmes and Missions (e.g. ESA Copernicus, NASA Landsat and PlanetScope to cite a few) opens the opportunity to collect satellite image time series (SITS) over the same study area to characterize and study the underlying spatio-temporal dynamics. Such rich information has been demonstrated to be largely beneficial in a variety of different fields, such as ecology~\citep{KoleckaGPPV18}, agriculture~\citep{IencoGDM17}, forestry~\citep{Wulder12}, environmental monitoring~\citep{HUANG2023397} and facilitating well-informed and sustainable decision-making policies~\citep{KAVVADA2020111930}. SITS are commonly exploited in conjunction with in situ ground truth data, acquired via costly and labor-intensive field campaigns, in order to derive timely up-to-date land cover (LC) maps over a specific region~\citep{GOMEZ201655}.

Typically, for the creation of LC maps over a region at a certain period of time, ground truth data are collected at a particular moment through expensive and time-demanding field campaigns. These data are then utilized in conjunction with SITS information through advanced machine learning algorithms~\citep{ZHONG2019430} to get the final LC map. While the access to high resolution EO data is no longer a major constraint, collecting up-to-date ground truth data constitutes a consumable (neither enduring nor lasting) effort. Once served its purpose, ground truth data will be disregarded loosing any further relevance. Furthermore, when the process is repeated (e.g., estimate agricultural production or potential biodiversity loss for a new year for the same or a related study site), new field campaigns must be afforded again with, in general, no way to profit from previous efforts.

Recently, endeavors related to the systematic and effective exploitation of available high-quality data have been increasing in both machine learning and computer vision communities. To this end, research actions in this direction have been proposed under the umbrella of data-centric Artificial Intelligence (AI)~\citep{abs-2303-10158}. Under this movement, the attention of researchers and practitioners is gradually shifting from advancing model design (model-centric AI) to enhancing the quality and quantity of the data (data-centric AI). 

Considering geospatial and EO data, the data-centric AI perspective is even more important since it can steer the community towards developing methodologies to provide further improvements related to the generalization ability with impact on real-world relevant problems and applications~\citep{abs-2312-05327}. Nevertheless, the two perspectives (model-centric and data-centric AI) play a complementary role in the larger machine learning deployment cycle since standard approaches still struggle to manage and exploit valuable data coming from different and heterogeneous distributions like, for instance, in the case of combining historical and up-to-date reference data for the downstream task of LC mapping~\citep{GaoYYQLZ23} where distribution shifts can be related to the different environmental and/or climatic factors that determine the acquisition conditions.

When data from different distributions are combined together under the same learning framework, challenges related to distribution shifts can impede the effective training of machine learning models~\citep{Ben-David10}. To cope with this issue, the most widely studied setting is Unsupervised Domain Adaptation (UDA)~\citep{WilsonC20} where the main goal is to learn a model over a labelled source domain and transfer it to an unlabelled target one. Recent UDA advances predominantly concentrate on deriving features that are domain-invariant. This is achieved through either aligning domains via data transformation~\citep{SunS16} or employing adversarial training~\citep{NEURIPS2018_ab88b157}, aiming to minimize the distribution gap between the source and target domains. In the context of remote sensing, early research have focused on devising UDA strategies tailored for high spatial resolution imagery~\citep{Tuia16}. Only recently, strategies have begun to emerge to analyse SITS data for spatial~\citep{NYBORG2022301} and temporal transfer tasks~\citep{CapliezIGBH23}.

Despite the efforts invested in designing and implementing UDA techniques, these approaches typically assume the complete absence of reference data for the target domain. However, for the LC mapping task, it is often reasonable to assume access to a certain amount of ground truth target data.
Moreover, the success of UDA depends largely on the discrepancy between the source and target distributions, making these methods susceptible to potential pitfalls and limited generalizability. In scenarios where a limited amount of labeled data is available for the target domain, the paradigm shifts to Semi-Supervised Domain Adaptation (SSDA). Current SSDA approaches generally aim to align the target data with the labeled source data with feature space mapping and self-training assignments using pseudo-labels~\citep{YuL23}.
Although the SSDA setting holds potential for various real-world problems, it remains largely unexplored when dealing with SITS data~\citep{LucasPSWP23} in the context of LC mapping.
Another related setting is Domain Generalization (DG)~\citep{JoY23,abs-2310-03007} wherein the objective is to learn classification models over a set of diverse labelled source domains that can generalize over new unseen target data. To the best of our literature survey, no DG frameworks have been proposed for EO data analysis. While DG and domain adaptation settings are closely related, DG operates inductively, assuming that target data are not available during the training stage. Conversely, in our problem setting, where the goal is to exploit together both historical and recent EO data along with reference data to enhance LC mapping, target data and their associated labels are already accessible.

In addressing the significant challenge outlined above, we present a novel framework, namely \method{}, rooted on recent advances in the field of domain adaptation/generalization. 
\method{} adopts a model-centric AI perspective, aiming to fulfill a data-centric AI objective related to the effective exploitation of EO and reference data coming from two different domains (e.g. historical data and recent ones) with the aim to give value again to historical and/or overlooked reference data, and enhance the accuracy of the recent LC mapping result. 

More precisely, \method{} is a deep learning framework built upon a pseudo-siamese network with unshared parameters that, given a sample, extracts simultaneously domain-invariant and domain-specific features, using the former to make the final decision. The objective is to disentangle useful information for the downstream LC mapping task while isolating and discarding domain specific features that can hinder the learning process in the presence of data belonging to multiple domains with unaligned data distributions. The disentanglement process is achieved by shaping a representation manifold, via contrastive learning, that jointly structures both domain-invariant and domain-specific features. In addition, \method{} integrates an effective supervision strategy~\citep{MOHAMMADI2023272} that further enforces the disentanglement process via multiple levels of supervision at different granularities.

Extensive experimental evaluations are carried out to assess the behavior of \method{} considering both baseline and domain adaptation/generalization approaches. To assess the behaviour of our method we perform both quantitative and qualitative evaluations considering two study sites covering extremely diverse and contrasted landscapes, namely \firstsite{} (located in the West-Africa region, in Burkina Faso) and \secondsite{} (located in centre Europe, France). For the former study site we consider a LC mapping task where data covering exactly the same geographical area are available for two different years (2020 and 2021) while, for the latter study site, we consider a crop type mapping task where data covering two closely related areas are available for the agricultural seasons 2018 and 2021.

This manuscript is organized as follows: 
Section~\ref{sec:method} introduces the proposed framework based on feature disentanglement and contrastive learning to enhance LC mapping combining multiple reference data. Study sites and the associated information are described in Section~\ref{sec:data}. The experimental evaluation and the related findings are reported and discussed in Section~\ref{sec:experiments}, while Section~\ref{sec:conclu} draws the conclusions of this paper.

\section{Proposed Deep Learning Framework} \label{sec:method}

\subsection{ Problem Formulation and Notations}
The proposed deep learning framework aims at improving the accuracy of LC mapping results obtained on recently acquired satellite data, i.e., target domain, by using pre-existing reference data coming from a different yet correlated domain, i.e., source domain. Typically, the source domain could consist of some readily-available historical or out-of-year data, for instance. Differences in climate, weather and other environmental conditions can lead to non-negligible distribution shifts within SITS data from the different domains. These shifts may prevent the full exploitation of the source data as a naive direct enrichment of the target data in a classic supervised learning setup  \citep[][]{bruzzone2009domain}. 
Moreover, in case the source data is more abundant, the learned classifier may likely be biased toward the source domain. Differently from the literature, the ultimate goal of REFeD is to maximize the classification performance in the target domain while taking full advantage of all the reference data available.

In this work, we suppose we are given a set of $N_t$ labeled samples from a target domain $\mathcal{D}^t= \{(x_i^t, y_i^t)\}_{i=1}^{N_t}$, for which we want to train a classifier. Moreover, we dispose of additional labelled data (say $N_s$ samples) from a source domain $\mathcal{D}^s= \{(x_i^s, y_i^s)\}_{i=1}^{N_s}$ that we aim to exploit in order to improve the performance of our classifier on the target domain. 

In our case, each sample $x_i \in \mathbb{R}^{T \times C}$ is the content of a pixel's $C$ spectral bands from a SITS defined over $T$ timestamps. In the considered experimental setup, we assume that the source and target SITS have the same number of $T$ timestamps. The corresponding label  $y_i \in \{1, \dots, K\}$ is given by one of $K$ existing classes, shared between source and target domains --- i.e., a closed-set scenario \citep{kundu2020}. Depending on the considered classification task, the classes could be, for instance, different crop and/or LC types. 
Let us also define as $y'_i \in \{ s, t\}$ the binary label associated to all the available labeled samples $\{(x_i, y'_i)\}_{i=1}^{N_s+N_t}$, which specify from which domain each spectral samples $x_i$ belongs, i.e., $\mathcal{D}^t$ or $\mathcal{D}^s$.

\subsection{REFeD: Overview}
\begin{figure}
    \centering
    \includegraphics[width=\textwidth]{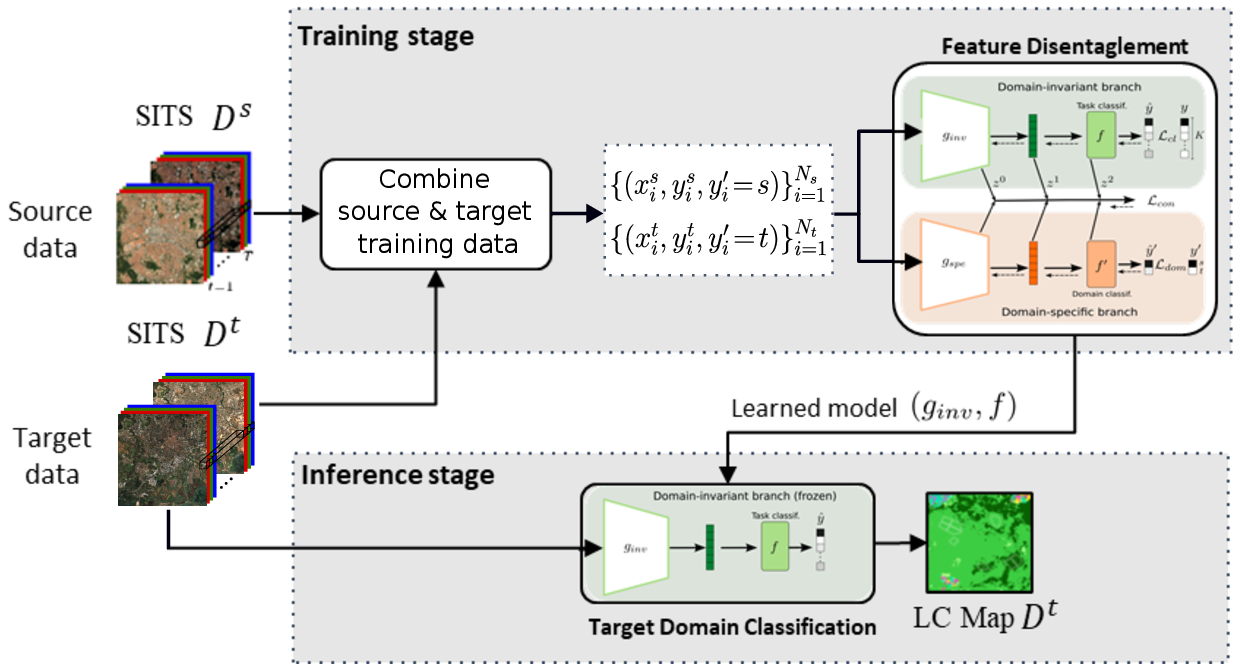}
    \caption{Overview of the proposed framework. Training and inference stages are distinguished: while the former is performed on data coming from both domains, the latter is done exclusively on target data and uses only the domain-invariant branch ($g_{inf}, f$) of the learned model.}
    \label{fig:overview}
\end{figure}

\Cref{fig:overview} shows an overview of the proposed deep learning framework, by depicting the data sources needed during the training and inference stages. In the first stage, the supervised classifier is trained using the reference data from both $\mathcal{D}^s$ and $\mathcal{D}^t$.
To take full advantage of reference data coming from distinct domains, we propose to disentangle the information carried by the labeled input data into two parts: 1) domain-specific information, and 2) domain-invariant information (i.e., useful discriminative information for the subsequent classification task). 
The former is closely related to the domain to which the data belong, thus potentially hindering the learning model's ability to generalize. The second contains semantic information associated with the underlying classes, thus usable knowledge that can be exploited later for the classification process. 

To this end, we take inspiration from current literature in domain generalization \citep{abs-2310-03007}, even though in their setting the target domain is considered to be unseen, i.e., they consider an inductive setting while in our case we have a transductive setup. 
In particular, in the training stage we leverage two branches with separate encoders to generate the feature vectors, i.e., $g_{spe}$ and $g_{inv}$. Dedicated losses, $\mathcal{L}_{cl}$, $\mathcal{L}_{dom}$ and $\mathcal{L}_{con}$ (detailed in \cref{ssec:disentangle})
are employed to effectively disentangle domain-specific from domain-invariant information in each of the two obtained embeddings, by training a domain classifier $f'(\cdot)$ and task classifier $f(\cdot)$, respectively. This condition allows us to benefit from the labeled samples available in all domains, taking into account the domain to which each labeled sample belongs.

To maximize the classification results obtained in the target domain, in the inference stage, the domain-specific encoder $g_{spe}$ is discarded and only the domain-invariant encoder $g_{inv}$ is considered.  In particular, we generate the LC map of the target domain using $g_{inv}$  for generating the feature representation to be classified along with the task classifier $f(\cdot)$ trained in the previous stage on the whole set of reference data. In the following, details are given.
\subsection{Feature disentanglement} \label{ssec:disentangle}
\begin{figure}
    \centering
    \includegraphics[width=0.85\textwidth]{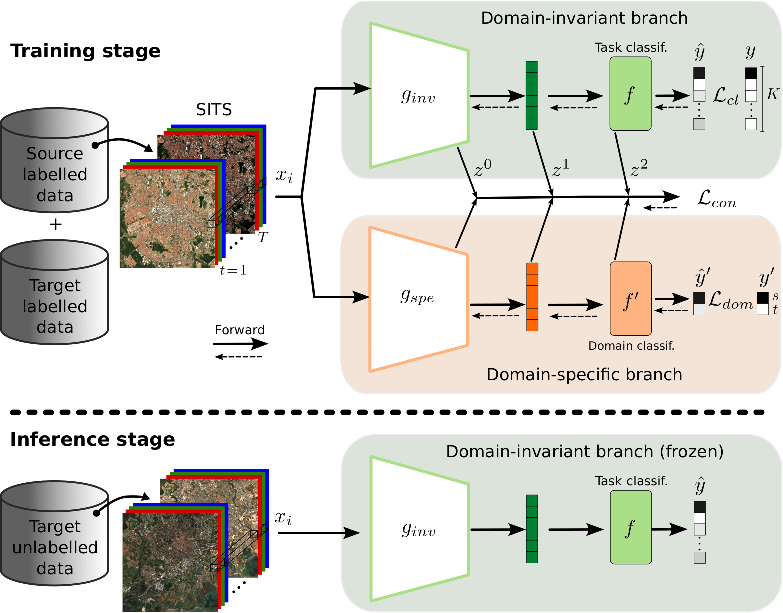}
    \caption{Architecture of the proposed pseudo-siamese network used in the training stage and composed of two independent branches which disentangle the domain-invariant information (top branch) from domain-specific information (bottom branch). Class ($\mathcal{L}_{cl}$) and domain ($\mathcal{L}_{dom}$)  discrimination losses used respectively on the top and bottom branches, while a multi-level contrastive loss ($\mathcal{L}_{con}$) is used to intermediate features at different depths from both branches. At inference time,  only the domain-invariant encoder is used for classifying the target domain.}
    \label{fig:architecture}
\end{figure}

Figure \ref{fig:architecture} depicts the pseudo-siamese network with unshared parameters used for feature disentanglement. The two encoders, denoted $g_{spe}$ and $g_{inv}: \mathbb{R}^{T \times C} \to \mathbb{R}^D$ for domain-specific and domain-invariant, respectively, share the same architecture but are learned independently with unshared weights via different loss functions.  

In the domain-invariant branch, a task classifier $f(\cdot)$ 
is applied to the domain-invariant features extracted by $g_{inv}$. In the domain-specific branch, the domain-specific features obtained via $g_{spe}$ are fed to a domain classifier $f'(\cdot)$ 
which encourages the domain-discriminant information to be channeled to this branch.

\paragraph{Domain classifier}
The domain classifier aims to accurately predict domain labels $y'_i \in \{ s, t\}$, i.e., determine if each sample $x_i$ belongs either to the source or the target domain, by minimizing a cross-entropy loss $\ell_{ce}$ as follows: 
\begin{align}
    \mathcal{L}_{dom} = \frac{1}{N_s+N_t}  \sum_{i=1}^{N_s+N_t} \ell_{ce}\left(f' \circ g_{spe}(x_i), y'_i\right) 
\end{align}

\paragraph{Task classifier}
In its turn, the task classifier $f: \mathbb{R}^D \to \{1, \dots, K\}$ maps the domain-invariant features onto one of the $K$ classes of interest guided by the following cross-entropy loss:
\begin{align}
    \mathcal{L}_{cl} = \frac{1}{N_s+N_t} \sum_{i=1}^{N_s+N_t} \ell_{ce}\left(f \circ g_{inv}(x_i), y_i\right) 
\end{align}

\paragraph{Contrastive learning}
To further decouple of the two separate branches, we employ contrastive loss which has shown promising results for feature disentanglement in previous works~\citep{abs-2310-03007}. In our case, since we have access to both the class labels and domain labels, we can employ the supervised version of the InfoNCE contrastive loss proposed by~\cite{KhoslaTWSTIMLK20}, where the positive pairs are given by all samples sharing the same label. However, here, the application is not straightforward since we have two separate label spaces (class and domain labels). To address this issue, we adopt a mixed label space 
$\mathcal{Y}_{mix}$ composed of $3K$ classes, where the domain-invariant features are mapped onto the $K$ first labels while the last $2K$ are reserved to the domain-specific features -- more specifically, $K$ for the source domain and $K$ for target domain. This leads to $\mathcal{Y}_{mix} = \{1, \dots, K, ~s1, \dots, sK, ~t1, \dots, tK\}$.%
\footnote{More generally, we define $\mathcal{Y}_{mix} = \mathcal{Y} \cup (\mathcal{Y} \times \mathcal{Y}')$ with a total of $|\mathcal{Y}_{mix}| \!=\! (|\mathcal{Y}'| \!+\! 1)|\mathcal{Y}|$ classes.}

Denoting $z$ the extracted embeddings (features) $g_{inv}(x)$ and $g_{spe}(x)$, we consider an augmented batch $I$ of size $2B$ containing both $g_{inv}(x_i)$ and $g_{spe}(x_i)$ features for each $i \in \{ 1, \dots, B\}$ in the original batch. The resulting supervised contrastive loss is defined as follows:
\begin{align} \label{eq:supervised_contrastive_loss}
    \mathcal{L}_{con} = -\sum_{i\in I} \frac{1}{|P(i)|} \sum_{p \in P(i)} \log \frac{\exp(z_i \cdot z_p / \tau)}{ \sum_{a \in I \setminus \{i\}} \exp(z_i \cdot z_a) / \tau}
\end{align}
where 
$P(i):= \{p \in  I \!\setminus\! \{ i\}: y_p = y_i\}$ with cardinality $|P(i)|$ is the set of \emph{positive} examples w.r.t. the current \emph{anchor} $i \in I := \{1, \dots, 2B\}$ and 
$\tau \in \mathbb{R}^{+}$ is a scalar temperature parameter. 
Therefore, the goal of this loss is to push together, in the feature space, the embeddings corresponding to the same category (positive examples) while repelling them from the rest (negative examples). The positive examples here correspond to those that share simultaneously: 1) the same class, among the K existing ones and 2) same domain type, among three options: source or target (for domain-specific features),  or domain-invariant.

\subsection{Multi-level supervision}

To further enforce the feature disentanglement, we propose to perform contrastive learning not only at the level of the encoder's output, but at multiple depths within the network architecture.

Then, the loss function described in \eqref{eq:supervised_contrastive_loss} is actually also applied to intermediate features at different depths of the network. For that matter, we denote $\mathcal{L}_{con}^l$ the contrastive loss \eqref{eq:supervised_contrastive_loss} applied to the intermediate features $z^l$ at depth $l$, as depicted in Figure \ref{fig:architecture}. Specifically, in our case, we use three levels of supervision with: $l=0$ for the encoder's last internal layer; $l=1$ for the encoder's output features; $l=2$ for the output of the classifier's first fully-connected layer. Note that $\mathcal{L}_{con}^l$ applies exclusively to features at depth $l$, which share the same space dimension, and, thus, features at different depths are never mixed together.

\subsection{Model summary and training}

The resulting loss is given by:
$$ \mathcal{L} = \mathcal{L}_{ce} + \mathcal{L}_{dom} +  {\textstyle \sum_l} \mathcal{L}^l_{con} $$

Empirically, we observed that weighting the different losses did not have a conclusive impact on the final model performance. For this reason, we used a simple unweighted sum of the three terms as described above. 

The proposed approach is completely agnostic to the encoder ($g_{inv}$, $g_{spe}$) and classifier modules architecture, but in this particular work we leverage the Temporal Convolutional Neural Network architecture (TempCNN) proposed by~\cite{PelletierWP19} for pixel-based SITS classification tasks.
The encoder part consists on three 1D-convolutional layers with 64 channels each. The classifier is composed of a fully-connected layer with 256 hidden units, batch normalisation and ReLU activation,  followed by a linear output layer with Softmax activation.

\section{Satellite Data and Ground Truth} 
\label{sec:data}
The first study site covers an area around the town of \textit{Koumbia}, in the Province of Tuy, \textit{Hauts-Bassins} region, in the south-west of Burkina Faso. This area has a surface of about 2\,338 $ km^2 $, and is situated in the sub-humid sudanian zone. The surface is covered mainly by natural savannah (herbaceous and shrubby) and forests, interleaved with a large portion of land (around 35\%) used for rainfed agricultural production (mostly smallholder farming). The main crops are cereals (maize, sorghum and millet) and cotton, followed by oleaginous and leguminous. Several temporary watercourses constitute the hydrographic network around the city of Koumbia.

\begin{figure}[!h]
\centering
\includegraphics[width=1.\textwidth]{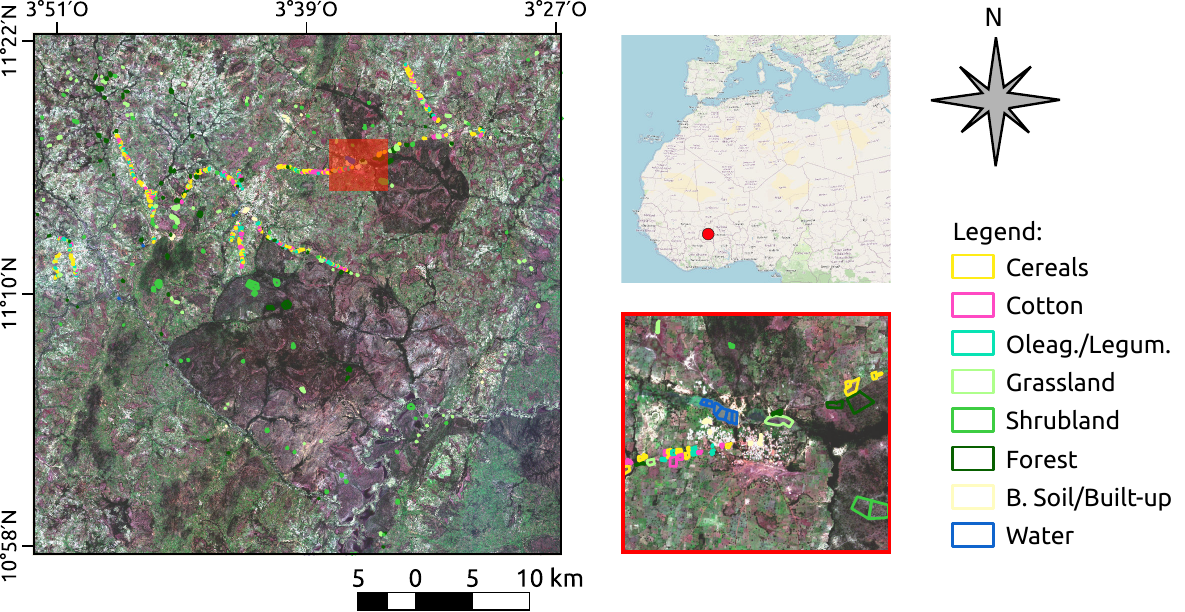}
\caption{View and location of Koumbia study site. The ground truth data coming from the 2020 year is superposed to a Sentinel-2 image covering the whole area. In the red box (bottom right) a more detailed view of the study site is depicted.}
\label{fig:koumbia_site}
\end{figure}

Figure~\ref{fig:koumbia_site} presents the study site with the 2020 reference data (ground truth) superposed on a Sentinel-2 image. A more detailed view corresponding to the red box in the overview is also depicted on the bottom right of the figure. A specific analysis of the ground truth is provided in Section~\ref{ssec:gt}.

The second study site covers two areas in the \textit{Centre Val de Loire} region located in the center of France. This region of France is characterized by an intensive agricultural activity with agricultural surfaces representing around 70\% of the whole region with cereals and oleaginous as major crops. The two areas have a cumulative surface of about 840 $ km^2 $.

\begin{figure}[!h]
\centering
\includegraphics[width=1.\textwidth]{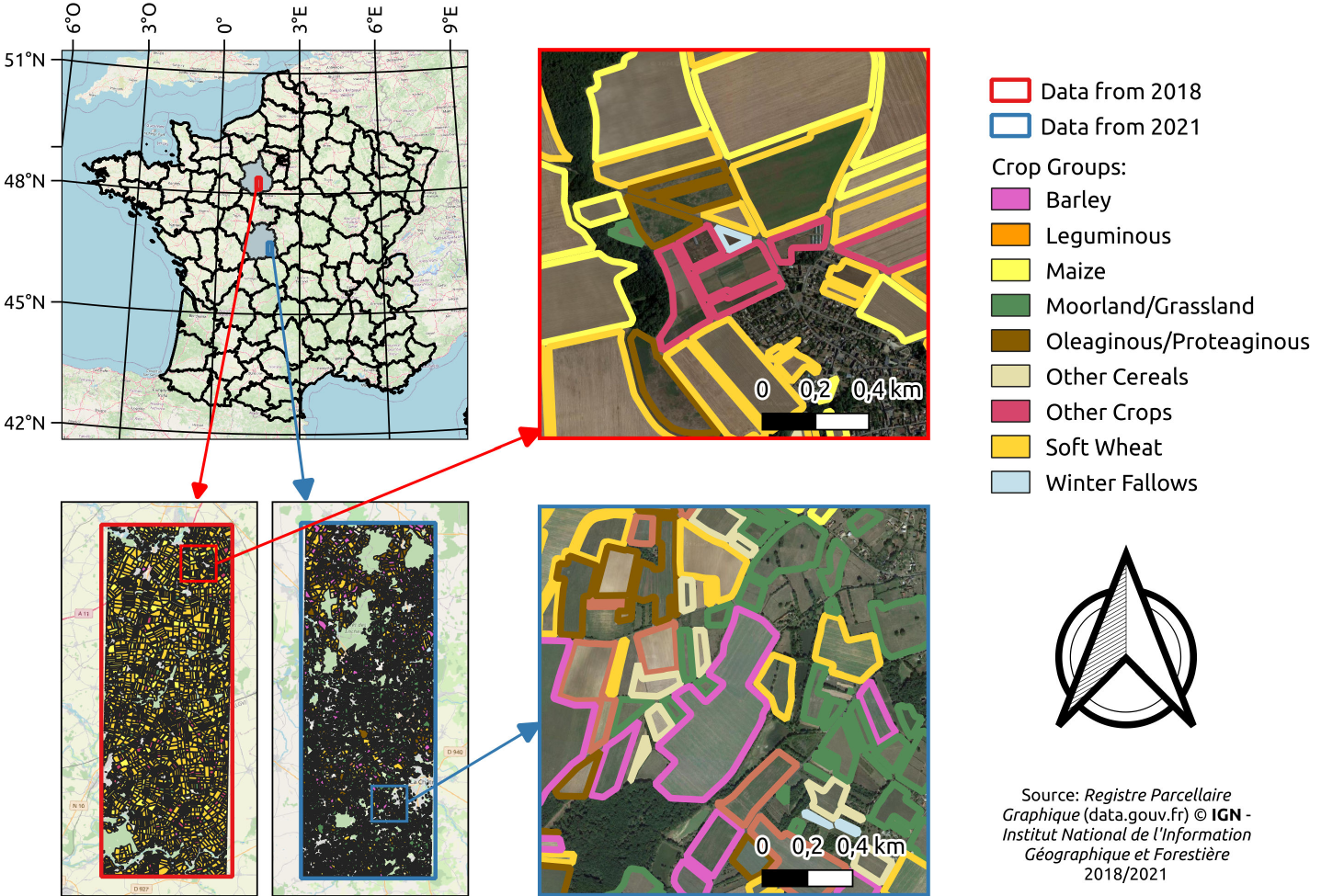}
\caption{View and location of the \secondsite{} study site. The ground truth data coming from the 2018 and 2021 year area superposed to a Sentinel-2 image. On the right, a detail for each of the areas is proposed, in red for 2018 and in blue for 2021.\label{fig:cvl}}
\end{figure}

Figure~\ref{fig:cvl} presents the two areas related to the \secondsite{} study site depicting reference (ground truth) data for year 2018 and 2021 superposed on a Sentinel-2 image. On the right of the figure, a detail for each of the areas is proposed, in red for 2018 and in blue for 2021. A specific analysis of the ground truth is provided in the Section~\ref{ssec:gt}.

\subsection{Satellite Image Time Series}
\label{ssec:sits}
For each study area and each considered year, we collected satellite image time series of Sentinel-2 imagery spanning an entire year via the Microsoft Planetary Computer platform~\footnote{\url{https://planetarycomputer.microsoft.com/}} that allows to access level-2A Sentinel-2 products. We consider all bands at 10~m and 20~m of spatial resolution for a total of 10 bands per image. We have conducted resampling of the SWIR 20~m bands to 10~m resolution, as well as image time series gap filling of cloudy pixels using multi-temporal linear interpolation as explained in~\citet{gapfilling} and gap-filled images were generated at a regular 10-day frequency resulting in a sequence of 72 images for each study area and year.

\subsection{Ground truth data}
\label{ssec:gt}

For the \firstsite{} study site, the ground truth data for 2020 and 2021 has been derived from a large agricultural LULC data set available online \citep{Jolivot2021}, mainly consisting of field data collected by local experts on several sites all over the tropics. For the \firstsite{} study site, the field surveys were conducted around the growing peak of the cropping season. The ground truth data cover the exact same surface for the two reference years.

For the \secondsite{} study site, the ground truth data for the first area, related to 2018, was obtained through the EuroCrop dataset~\citep{schneider_eurocrops_2023} while the ground truth data for the second area, related to 2021, are gathered from the RPG (\textit{Registre Parcellaire Graphique}), the French land parcel identification system. The data covers only agricultural areas in order to set up a crop type mapping task. This second dataset covers a problem that implies both spatial and temporal transfer at the same time.


For both study sites, ground truth has been assembled in a Geographic Information System (GIS) vector file, containing a collection of polygons, each attributed with a LC or crop type category based on information reported in the original database -- see Tables~\ref{tab:gt} and \ref{tab:gt2} for statistics about ground truth data distribution.

\begin{table}[!ht]
\caption{Ground truth statistics for year 2020 and 2021 on the \firstsite{} study site. \label{tab:gt}}
\centering
\small
\begin{tabular}{lc|cc|cc} \hline
 Class Name & Class ID. & \multicolumn{2}{|c|}{2020} & \multicolumn{2}{|c}{2021} \\ 
\cmidrule(lr){1-1} \cmidrule(lr){2-2} \cmidrule(lr){3-4} \cmidrule(lr){5-6}
 & & \# Polygons & \# Pixels & \# Polygons & \# Pixels \\  
\cmidrule(lr){1-1} \cmidrule(lr){2-2} \cmidrule(lr){3-3} \cmidrule(lr){4-4} \cmidrule(lr){5-5} \cmidrule(lr){6-6} 
    Cereals & 1 & 230 & 9731 & 268 & 11435 \\
    Cotton & 2 & 139 & 6971 & 121 & 6575 \\
    Oleaginous / Leguminous & 3 & 281 & 7950 & 263 & 7316 \\ 
    Grassland & 4 & 122 & 12998 & 113 & 11100 \\  
    Shrubland & 5 & 83 & 22546 & 90 & 24324 \\
    Forest & 6 & 82 & 17435 & 82 & 16984 \\
    Bare soil / Built-up & 7 & 51 & 1125 & 51 & 1022 \\
    Water & 8 & 10 & 1205 & 10 & 1205 \\
 \cmidrule(lr){1-1} \cmidrule(lr){2-2} \cmidrule(lr){3-3} \cmidrule(lr){4-4} \cmidrule(lr){5-5} \cmidrule(lr){6-6} 
 & Total & 998 & 79961 & 998 & 79961 \\
 \bottomrule
\end{tabular}
\end{table}

\begin{table}[!ht]
\caption{Ground truth statistics for year 2018 and 2021 on the \secondsite{} study site. \label{tab:gt2}}
\centering
\small
\begin{tabular}{lc|cc|cc} \hline
 Class Name & Class ID. & \multicolumn{2}{|c|}{2018} & \multicolumn{2}{|c}{2021} \\ 
\cmidrule(lr){1-1} \cmidrule(lr){2-2} \cmidrule(lr){3-4} \cmidrule(lr){5-6}
 & & \# Polygons & \# Pixels & \# Polygons & \# Pixels \\  
\cmidrule(lr){1-1} \cmidrule(lr){2-2} \cmidrule(lr){3-3} \cmidrule(lr){4-4} \cmidrule(lr){5-5} \cmidrule(lr){6-6} 
    Soft wheat & 1 & 1048 & 9388 & 1341 & 9268\\
    Maize & 2 & 198 & 9442 & 264 & 9437\\
    Barley & 3 & 674 & 9345 & 527 & 9222\\  
    Other cereals & 4 & 421 & 9288 & 360 & 9473\\  
    Oleaginous / Proteaginous & 5 & 842 & 9222 & 775 & 9168\\
    Winter Fallows & 6 & 413 & 8390 & 611 & 7396\\
    Leguminous & 7 & 5 & 3418 & 12 & 4753\\
    Fodder & 8 & 255 & 9350 & 105 & 7422\\
    Meadow & 9 & 4132 & 8254 & 127 & 6030\\
    Other crops & 10 & 184 & 8183 & 355 & 9681\\ 
 \cmidrule(lr){1-1} \cmidrule(lr){2-2} \cmidrule(lr){3-3} \cmidrule(lr){4-4} \cmidrule(lr){5-5} \cmidrule(lr){6-6} 
 & Total & 8172 & 84280 & 12649 & 81850\\ \hline
\end{tabular}
\end{table}

\section{Experiments}
\label{sec:experiments}
In this section, we report and discuss the experimental evaluation carried out on the study sites presented in Section~\ref{sec:data}. 
Our objective is to evaluate the performance of \method{} across various dimensions. First, we undertake a quantitative assessment comparing the performance of \method{} against baselines and competing approaches. Second, we conduct a qualitative examination of the land cover maps generated by \method{}. Lastly, we inspect the internal representations learned by our model, visually comparing them to those of some of the top-performing competitors.

\subsection{Competing methods}
\label{ssec:expe-methods}
With the goal to assess the performance of \method{} w.r.t. baselines and strategies coming from SSDA and DG literature, we consider:
\begin{itemize}
\item \textit{Only Source}: this strategy trains a model only considering source data and, then, the obtained classifier is directly deployed on the target data. The main purpose of this method is to have an empirical estimate about the distribution shift between source and target domains. We implement this baseline considering both Random Forest and TempCNN~\citep{PelletierWP19} as classifiers.
\item \textit{Only Target}: this strategy trains a model only considering the target labelled data and, then, the obtained classifier is employed to classify the remaining target samples. The main purpose of this method is to provide the reference performances without the use of historical or out-of-year data. We implement this baseline considering both Random Forest and TempCNN~\citep{PelletierWP19} as classifiers.
\item \textit{Source+Target}: this strategy trains a model over both source and target labelled data. Then, the obtained classifier is employed to classify the remaining target samples. Here, the data coming from different distributions are mixed together constituting a multi-domain training dataset. We implement this baseline considering both Random Forest and TempCNN~\citep{PelletierWP19} as classifiers.
\item \textit{Fine Tuning}: this strategy trains a model over the labelled source domain and, then, the resulting model is fine tuned on target labelled samples. This is an alternative way to combine both source and target data.  We implement this baseline considering only the TempCNN~\citep{PelletierWP19} approach.
\item \textit{Sourcerer}~\citep{LucasPSWP23}: this recent SSDA approach has been proposed to cope with the analysis of SITS data for the downstream task of LULC mapping. Sourcerer is a bayesian-inspired, deep learning-based framework, that internally exploits the TempCNN model as backbone, similarly to \method. The technique leverages a deep learning model trained on a source domain and then fine-tunes the model on the available target domain via a regularizing term that automatically adjusts the degree to which the model weights are modified to fit the target data.
\item \textit{POEM}~\citep{JoY23}: this recent DG approach learns domain-invariant and domain-specific representations via a pseudo-siamese network with unshared weights and it enforces polarization via orthogonality constraints. This approach was primarily introduced for image classification. In order to transfer it on SITS data, also for this competitor we employ the TempCNN model as a backbone.
\end{itemize}

\subsection{Experimental settings}
\label{ssec:expe-settings}
For all the competing approaches, labelled source data are entirely employed while, for the target domain, data are split into three parts: training, validation and test sets following a proportion of 50\%, 20\% and 30\% of the original target data set, respectively. Regarding the \textit{Only Source} baseline, the model is trained considering only the labelled source data.
Furthermore, with the aim to avoid possible spatial bias in the evaluation procedure~\citep{karasiak_ml_2021}, we impose that all the pixels belonging to the same object will be exclusively associated to one of the data partitions (training, validation or test). The splitting procedure is repeated five times and the average results are reported.

Concerning the evaluation tasks, according to the data presented in Section~\ref{sec:data}, we set up two transfer tasks per benchmark. Each transfer task is denoted as ($\mathcal{D}_s$ + $\mathcal{D}_t$ $\rightarrow$ $\mathcal{D}_t$) where the right arrow indicates the transfer direction from the combined source/target labelled training dataset ($\mathcal{D}_s$ + $\mathcal{D}_t$) to the test target ($\mathcal{D}_t$) dataset. For the \firstsite{} study site, we consider as transfer tasks (2020 + 2021 $\rightarrow$ 2021) and (2021 + 2020 $\rightarrow$ 2020) and for the \secondsite{} study site, we consider the transfer tasks (2018 + 2021 $\rightarrow$ 2021) and (2021 + 2018 $\rightarrow$ 2018).

The values of the SITS benchmarks were scaled per year and per band considering the 2nd and 98th percentile of the data distribution as minimum and maximum values. The assessment of the model performances was done considering the following metrics: \textit{Weighted F1-score} (simply indicated with \textit{F1-score}) and \textit{Accuracy} (global precision). 

\textbf{Implementation details:} For the neural network approaches, the training stage has been conducted for 200 epochs. For methods based on fine-tuning, we used 100 epochs for the initial training and 100 epochs for the fine-tuning stage. For all methods, we adopt a learning rate of 10$^{-4}$, the AdamW~\citep{LoshchilovH19} optimizer and a batch size of 256. Regarding \method{}, based on recent literature on contrastive learning~\citep{ChenZXCD0TZC22}, we set the temperature hyperparameter $\tau$ to 0.07 and we consider a batch size of 512 since it has been noted that contrastive loss benefits from larger batch sizes. The drop out value is set to 50\%. Considering Random Forest classifiers, we optimize the model via the tuning of one parameter: the number of trees in the forest. We vary this parameter in the range \{100, 200, 300, 400, 500\}. The optimization of this parameter is based on the validation set.

Experiments are carried out on a workstation with a dual Intel (R) Xeon (R) CPU E5-2667v4 (@3.20GHz) with 256 GB of RAM and four TITAN X (Pascal) GPU. All the deep learning methods are implemented using the Pytorch deep learning library. All the models run on a single GPU. The Random Forest is implemented using the Python \textit{Scikit-learn} library~\citep{scikit-learn}. 

\subsection{Quantitative results}
\label{ssec:expe-quant-results}

\begin{table*}[]
    \centering
    \begin{tabular}{c|c|c|c|c|c} \hline
        \textbf{Strategy} & \textbf{Method} & \multicolumn{2}{|c|}{2020 + 2021 $\rightarrow$ 2021} & \multicolumn{2}{|c}{2021 + 2020 $\rightarrow$ 2020} \\ \hline
        & & \textbf{F1-Score} & \textbf{Accuracy} & \textbf{F1-Score} & \textbf{Accuracy} \\ \hline
        \multirow{2}{*}{\textit{Only Source}} 
            & RF & 72.40 $\pm$ 3.94 & 72.25 $\pm$ 4.08 & 71.98 $\pm$ 3.54 & 71.81 $\pm$ 3.63 \\
            & TempCNN & 64.54 $\pm$ 4.32 & 65.25 $\pm$ 4.14 & 69.54 $\pm$ 4.88 & 69.95 $\pm$ 5.00  \\ \hline
        \multirow{2}{*}{\textit{Target domain}} 
            & RF & 77.56 $\pm$ 2.82 & 77.45 $\pm$ 2.90 & 76.78 $\pm$ 3.53 & 76.65 $\pm$ 3.54\\
            & TempCNN & 76.59 $\pm$ 2.94 & 76.63 $\pm$ 2.91 & 75.54 $\pm$ 5.04 & 75.47 $\pm$ 5.09 \\ \hline
        \multirow{2}{*}{\textit{Source+Target}} 
            & RF & 77.49 $\pm$ 3.90 & 77.30 $\pm$ 3.93 & 78.42 $\pm$ 4.02 & 78.32 $\pm$ 4.03 \\
            & TempCNN & \underline{78.60} $\pm$ 2.94 & \underline{78.48} $\pm$ 3.03 & \underline{78.95} $\pm$ 4.27 & \underline{78.92} $\pm$ 4.27  \\ \hline
        \textit{Fine Tuning} & TempCNN & 78.04 $\pm$ 2.75 &  78.00 $\pm$ 2.78 & 78.56 $\pm$ 3.63 & 78.55 $\pm$ 3.64 \\ \hline
        \textit{} & Sourcerer & 76.72 $\pm$ 2.63 & 76.54 $\pm$ 2.64 & 76.34 $\pm$ 3.94  & 76.24 $\pm$ 3.99 \\ \hline        
        \textit{} & POEM & 78.35 $\pm$ 3.43 & 78.27 $\pm$ 3.41 & 78.41 $\pm$ 4.63 & 78.35 $\pm$ 4.66 \\  \hline
        \textit{} & \method{} & \textbf{79.23} $\pm$ 3.33 & \textbf{79.17} $\pm$ 3.27 & \textbf{82.15} $\pm$ 3.65 & \textbf{82.09} $\pm$ 3.67\\ 
        \hline
    \end{tabular}
    \caption{Overall performances (F1 scores) on Koumbia \label{tab:F1-Koumbia} }    
\end{table*}

\begin{table*}[]
    \centering
    \begin{tabular}{c|c|c|c|c|c} \hline
        \textbf{Strategy} & \textbf{Method} & \multicolumn{2}{|c|}{2018 + 2021 $\rightarrow$ 2021} & \multicolumn{2}{|c}{2021 + 2018 $\rightarrow$ 2018} \\ \hline
        & & \textbf{F1-Score} & \textbf{Accuracy} & \textbf{F1-Score} & \textbf{Accuracy} \\ \hline
        \multirow{2}{*}{\textit{Only Source}} 
            & RF &  69.66 $\pm$ 1.64 & 72.35 $\pm$ 1.07 & 59.31 $\pm$ 3.19  & 62.36 $\pm$ 2.56\\
            & TempCNN & 47.96 $\pm$ 1.23 & 46.26 $\pm$ 0.76 & 43.10 $\pm$ 2.46 & 41.68 $\pm$ 3.31\\ \hline
        \multirow{2}{*}{\textit{Target domain}} 
            & RF & 79.95 $\pm$ 2.45 & 80.70 $\pm$ 1.78 & 70.63 $\pm$ 3.05 & 71.86 $\pm$ 2.55\\
            & TempCNN & 82.60 $\pm$ 1.15  & 80.61 $\pm$ 1.41 & 73.39 $\pm$ 3.12 & 71.46 $\pm$ 3.46\\ \hline
        \multirow{2}{*}{\textit{Source+Target}} 
            & RF & 79.00 $\pm$ 2.24 & 79.83 $\pm$ 1.60 & 68.94 $\pm$ 2.66 & 70.45 $\pm$ 2.31\\
            & TempCNN & \underline{83.46} $\pm$ 1.40 &  81.54 $\pm$ 2.07 & \underline{75.17} $\pm$ 3.61 & 74.17 $\pm$ 3.71 \\ \hline
        \textit{Fine Tuning} & TempCNN & 82.94 $\pm$ 2.04 & \underline{83.51} $\pm$ 1.75 & 74.43 $\pm$ 3.78 & 75.70 $\pm$ 3.29\\ \hline
        \textit{} & Sourcerer & 82.76 $\pm$ 1.69 & 83.49 $\pm$ 1.29 & 74.02 $\pm$ 3.36 & 74.93 $\pm$ 2.90 \\ \hline        
        \textit{} & POEM & 82.89 $\pm$ 1.71 & \underline{83.51} $\pm$ 1.31 &  74.28 $\pm$ 3.11 & \underline{75.14} $\pm$ 2.80 \\  \hline
        \textit{} & \method{} & \textbf{84.45} $\pm$ 1.61 & \textbf{84.86} $\pm$ 1.35 & \textbf{77.60} $\pm$ 3.15 & \textbf{78.02} $\pm$ 2.90\\ 
        \hline
    \end{tabular}
    \caption{Overall performances (F1 scores) on \secondsite{} (CVL) \label{tab:F1-CVL} }    
\end{table*}

\Cref{tab:F1-Koumbia,tab:F1-CVL} summarize the results obtained for the two study areas, \firstsite{} and \secondsite{} respectively by reporting the average F1-score and the Accuracy considering the different combination of methods and strategies.  As expected, regardless of the dataset, for both the RF and TempCNN classifiers, the lowest accuracy is obtained when only source-labeled data are considered, while the highest classification results are obtained when training data from both domains are used.  

By focusing our attention on the TempCNN architecture, the F1-scores obtained using only the source-labeled data are 64.54 and 69.54 in  \firstsite{} on EO data acquired in 2021 and 2020, respectively, and 47.96 and 43.10 in  \secondsite{} on EO data acquired in 2021 and 2018, respectively. Although valuable, the historical reference data may not be completely representative of the recently acquired EO data. Moreover, the class statistical distributions of SITS acquired over different years can severely shift. Using only the target-labeled data the obtained accuracy increases, i.e., F1 scores of 76.59 and 75.54 in \firstsite{} on EO data acquired in 2021 and 2020, respectively, and 82.60 and 73.39 in  \secondsite{} on EO data acquired in 2021 and 2018, respectively. It is worth noting that the importance of using labeled data from the target domain is even more visible in the \secondsite{} dataset since the source and target domains are different from both the spatial and temporal viewpoints.  The joint use of source and target labeled data has a positive impact on the classification performances, leading to F1 scores of 78.60 and 78.95 in \firstsite{} on EO data acquired in 2021 and 2020, respectively, and 83.46 and 75.17 in \secondsite{} on EO data acquired in 2021 and 2018, respectively. 

These results further improve when using SSDA and DG methods, which better combine source and target information. However, the highest classification accuracies are obtained by the proposed approach \method{} which achieves F1 scores of 79.23 and 82.15 in  \firstsite{} on EO data acquired in 2021 and 2020, respectively, and 84.45 and 77.60  in \secondsite{} on EO data acquired in 2021 and 2018, respectively.


\subsubsection{Per-class analysis}
\label{ssec:per-class-analysis}

Per-class performances are detailed in \Cref{tab:F1-Koumbia-perclass1,tab:F1-Koumbia-perclass2} for the \firstsite{} study site and in \Cref{tab:F1-CVL-perclass1,tab:F1-CVL-perclass2} and \secondsite{} respectively. 

Concerning the \firstsite{} benchmark, \Cref{tab:F1-Koumbia-perclass1,tab:F1-Koumbia-perclass2}, we can observe that, no matter the transfer task, \method{} achieves the best performances in terms of F1-Score on all the agricultural land cover classes. This is of particular interest since such classes are characterized by strong shifts from one cultural year to another one due to crop rotations related to the underlying agricultural practices. A notable improvement, related to the proposed method, can also be noted on the \textit{Forest} land cover class, especially for the transfer task (2021 + 2020 $\rightarrow$ 2020). Regarding all the other land cover classes, \method{} achieves comparable results w.r.t. to all the other competing methods with, for all these cases, less than a point of difference in terms of F1-score.

Regarding the \secondsite{} benchmark, \Cref{tab:F1-CVL-perclass1,tab:F1-CVL-perclass2}, we can note that, \method{} obtains a systematic improvement, in terms of F1-Score, for the family of cereal classes (\textit{Soft wheat}, \textit{Maize}, \textit{Barley} and \textit{Other cereals}), regardless of the transfer task. Notably, the most significant enhancement is observed in the \textit{Leguminous} crop class, where \method{} attains an improvement between 6 and 10 points of F1-Score compared to the best competing approach. This is even interesting since the \textit{Leguminous} class is the most underrepresented crop type in the considered benchmark in terms of number of samples. This point further underscores the quality of the proposed approach demonstrating its ability to handle scenarios characterized by significant class imbalances, a common situation in real-world applications.
Finally, still regarding the \secondsite{} benchmark, we can underline that for the transfer task (2021 + 2018 $\rightarrow$ 2018), \method{} achieves the best performances for all the crop type classes.


\begin{table}[]
\centering
\begin{tabular}{l|c|c|c|c} \hline
\textbf{Class} & \textbf{TempCNN}  & \textbf{Sourcerer} & \textbf{POEM} & \textbf{REFeD} \\
& (Source+Target) &  & & \\ \hline
Cereals & 74.06 & 70.79  & 72.48 & \textbf{75.73} \\
Cotton & 75.77 & 72.79 & 76.04 &  \textbf{77.87} \\
Oleaginous & 64.18 &  60.22 & 64.24 & \textbf{66.96} \\ 
Grassland & \textbf{78.58} & 75.62  & 78.11 &  77.61 \\  
Shrubland & \textbf{78.90} &  76.47 & 78.32 &  77.90 \\
Forest & 84.27 & 85.06 & 84.78 & \textbf{86.05} \\
Bare soil & \textbf{83.45} & 80.11 & 82.83 & 83.04 \\
Water & \textbf{100.0} & \textbf{100.0} & \textbf{100.0} &  \textbf{100.0} \\
\hline    
\end{tabular}
    \caption{Per-class average F1 scores for \firstsite{} scenario (2020 + 2021 $\rightarrow$ 2021).}
    \label{tab:F1-Koumbia-perclass1}
\end{table}

\begin{table}[]
\centering
\begin{tabular}{l|c|c|c|c} \hline
\textbf{Class} & \textbf{TempCNN}  & \textbf{Sourcerer} & \textbf{POEM} & \textbf{REFeD} \\
& (Source+Target) &  & & \\ \hline
Cereals & 77.48  & 74.25 & 76.39 & \textbf{79.38} \\
Cotton & 78.10 & 75.48  & 77.18 & \textbf{82.96} \\
Oleaginous & 72.91 & 69.93 & 71.69 & \textbf{75.14} \\ 
Grassland & 85.22 & 82.68 & 84.14 & \textbf{86.38} \\  
Shrubland & 77.31 & 74.64 & 77.29 & \textbf{81.38} \\
Forest & 77.03 & 74.97  & 77.33 & \textbf{81.94} \\
Bare soil & \textbf{79.55} & 77.61 & 76.78 & 78.71 \\
Water & \textbf{100.0} & \textbf{100.0} & \textbf{100.0} & \textbf{100.0} \\
\hline    
\end{tabular}
    \caption{Per-class average F1 scores for \firstsite{} scenario (2021 + 2020 $\rightarrow$ 2020).}
    \label{tab:F1-Koumbia-perclass2}
\end{table}

\begin{table}[]
\centering
\begin{tabular}{l|c|c|c|c} \hline
\textbf{Class} & \textbf{TempCNN}  & \textbf{Sourcerer} & \textbf{POEM} & \textbf{REFeD} \\
& (Source+Target) &  & & \\ \hline
Soft wheat & 91.35 & 91.53 & 90.52 & \textbf{91.91} \\
Maize & 94.91 & 92.72 & 94.73 & \textbf{96.14} \\
Barley & 94.94 & 94.71 & 94.75 & \textbf{95.58} \\ 
Other cereals & 85.55 & 85.84 & 84.36 & \textbf{87.54} \\
Oleaginous & 86.76 & \textbf{88.56} & 86.87 & 86.43 \\
Winter Fallows & 73.82 & 71.98 & 73.22 & \textbf{73.87} \\
Leguminous & 55.26 & 52.26 & 54.93 & \textbf{61.63}\\
Fodder & 74.31 & 73.04 & 72.11 & \textbf{75.82}\\
Meadow & \textbf{74.35} & 71.33 & 73.92 & 73.48 \\
Other crops & \textbf{84.16} & 83.88 & 84.06 & 83.92 \\
\hline    
\end{tabular}
    \caption{Per-class average F1 scores for \secondsite{} scenario (2018 + 2021 $\rightarrow$ 2021).}
    \label{tab:F1-CVL-perclass1}
\end{table}

\begin{table}[]
\centering
\begin{tabular}{l|c|c|c|c} \hline
\textbf{Class} & \textbf{TempCNN}  & \textbf{Sourcerer} & \textbf{POEM} & \textbf{REFeD} \\
& (Source+Target) &  & & \\ \hline
Soft wheat & 85.55 & 85.23 & 84.21 & \textbf{85.80} \\
Maize & 85.65 &  85.48 & 86.23 & \textbf{87.04} \\
Barley & 94.28 &  94.34 & 93.63 & \textbf{94.63} \\ 
Other cereals & 73.35 & 73.87 & 71.35 & \textbf{73.79}\\
Oleaginous & 72.49 & 70.01 & 73.02 & \textbf{76.98} \\
Winter Fallows & 74.79 & 74.24  & 73.78 & \textbf{75.13} \\
Leguminous & 54.77 & 46.51 & 53.05 & \textbf{65.15}\\
Fodder & 70.47 & 68.24 & 66.91 & \textbf{72.07}\\
Meadow & 72.99 & 71.62 & 71.34 & \textbf{73.04} \\
Other crops & 57.40 & 57.76 & 57.66 & \textbf{64.06} \\
\hline    
\end{tabular}
    \caption{Per-class average F1 scores for \secondsite{} scenario (2021 + 2018 $\rightarrow$ 2018).}
    \label{tab:F1-CVL-perclass2}
\end{table}

\subsection{Visual analysis}
\label{ssec:vis-analysis}
In this part of the experimental assessment, we provide qualitative analyses to further evaluate the behaviour of \method{} considering the \firstsite{} site, in the transfer task (2021 + 2020 $\rightarrow$ 2020). To this end, in addition to our framework, we also consider the top-performing competitors: \textit{POEM}, \textit{Sourcerer} and \textit{TempCNN} (Source + Target). We first inspect some extracts from the obtained land cover maps and, then we visually examine the internal representations learned by the different methods by means of the t-SNE~\citep{Maaten2008VisualizingDU} dimensionality reduction technique.

\subsubsection{Land cover maps}
\label{ssec:lcm}

\begin{figure}
    \centering
    \def\arraystretch{0}
    \setlength\tabcolsep{0pt}
    \begin{tabular}{ccccc}
         \footnotesize
         \textit{SPOT6/7 Image} & \textit{TempCNN} & \textit{Sourcerer} & \textit{POEM} & \textit{\textbf{REFeD}} \\
         \includegraphics[width=.2\columnwidth]{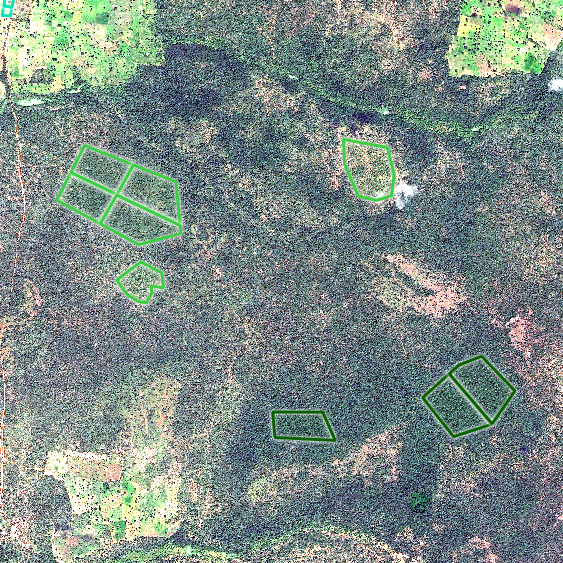} & 
         \includegraphics[width=.2\columnwidth]{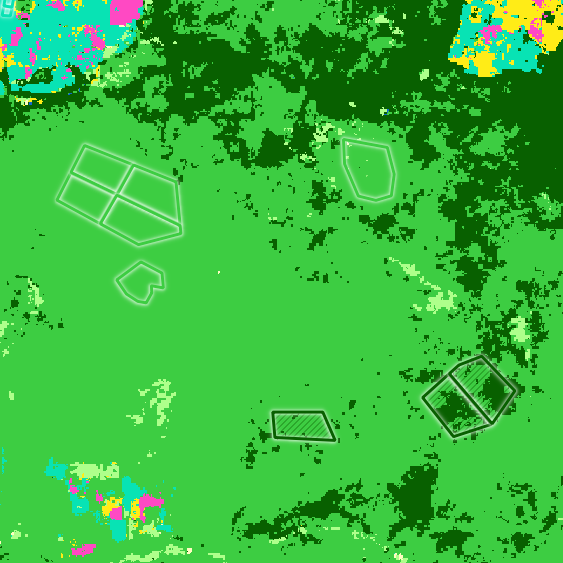} &
         \includegraphics[width=.2\columnwidth]{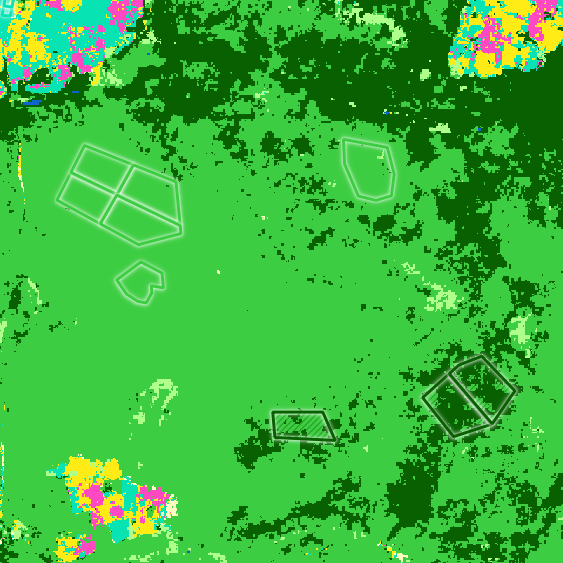} &
         \includegraphics[width=.2\columnwidth]{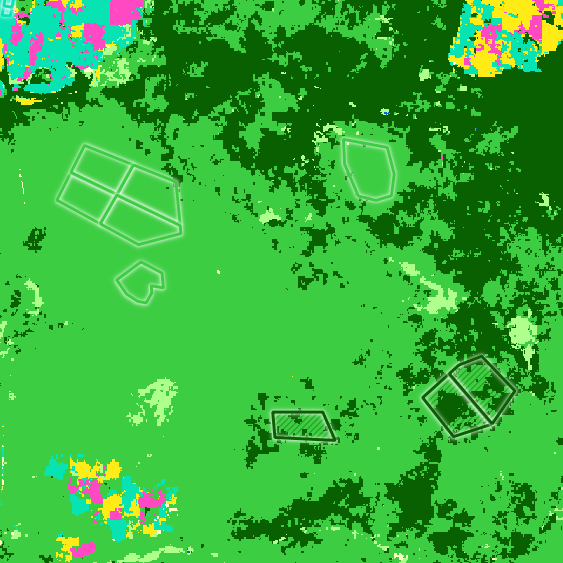} &
         \includegraphics[width=.2\columnwidth]{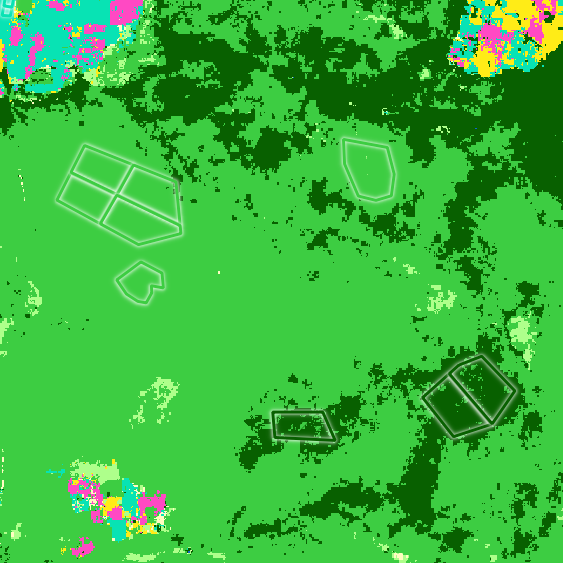} \\
         \includegraphics[width=.2\columnwidth]{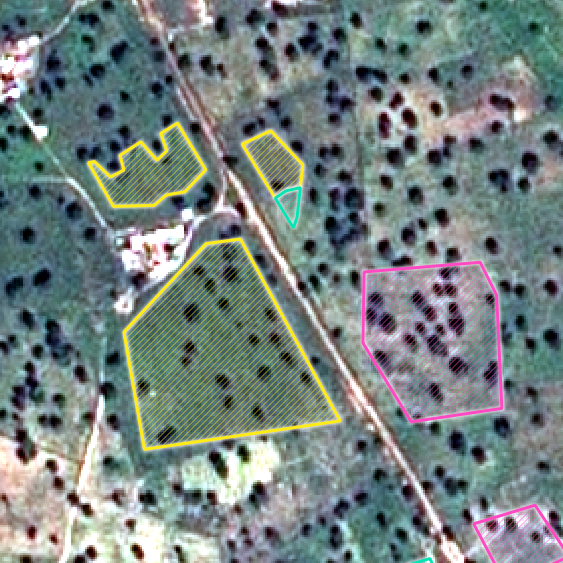} & 
         \includegraphics[width=.2\columnwidth]{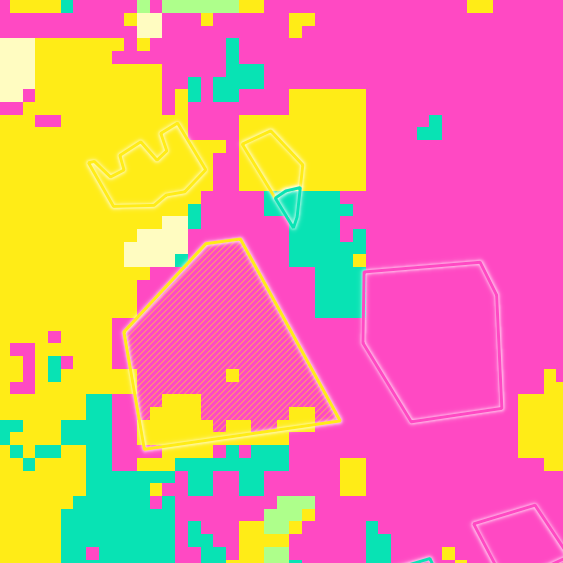} &
         \includegraphics[width=.2\columnwidth]{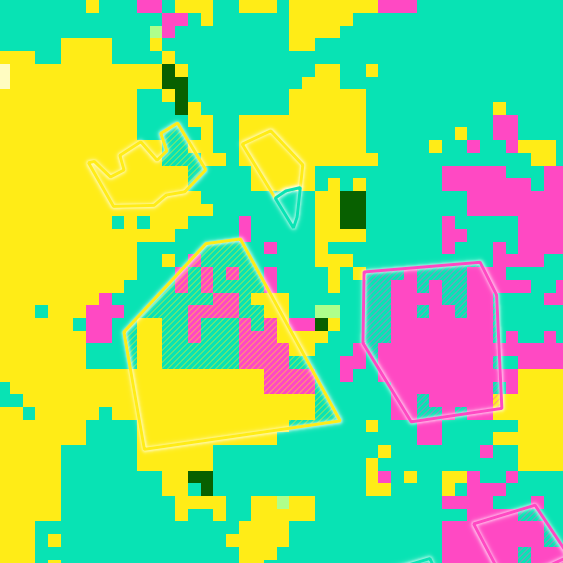} &
         \includegraphics[width=.2\columnwidth]{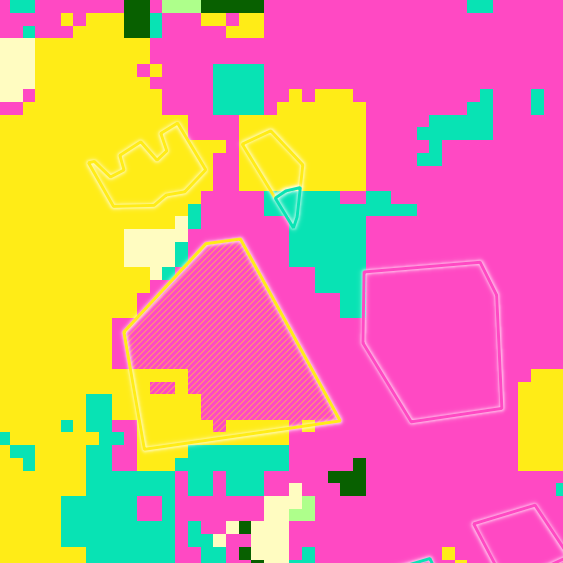} &
         \includegraphics[width=.2\columnwidth]{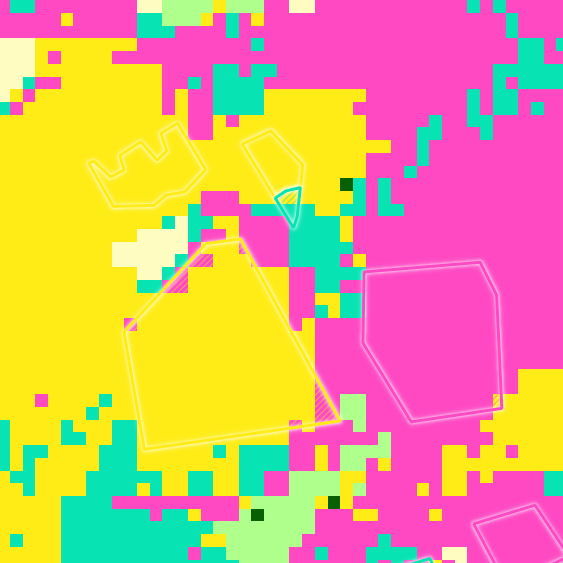} \\
         \includegraphics[width=.2\columnwidth]{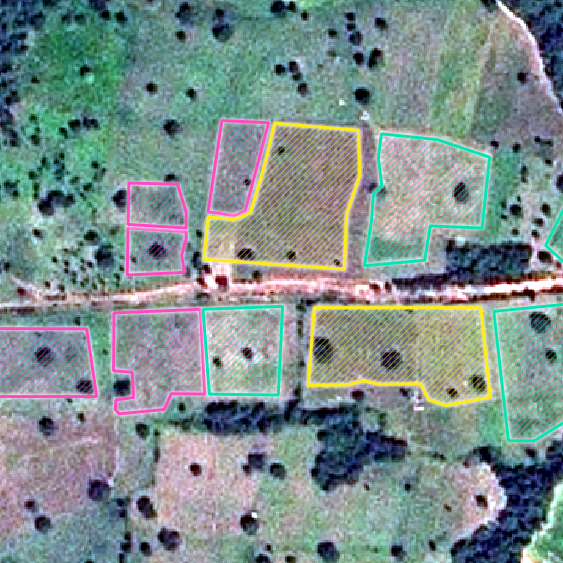} & 
         \includegraphics[width=.2\columnwidth]{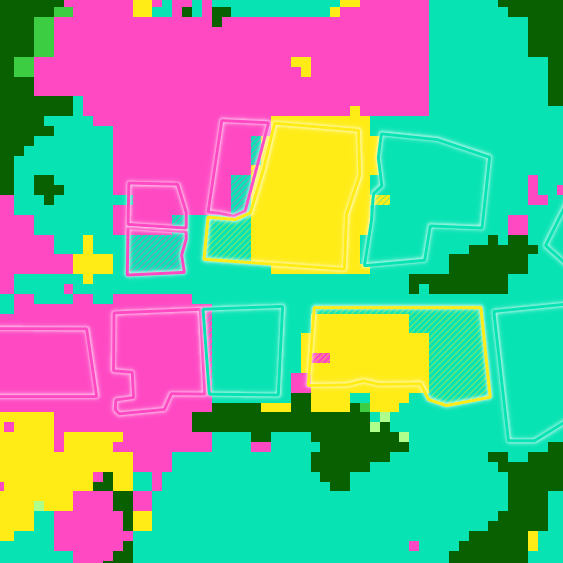} &
         \includegraphics[width=.2\columnwidth]{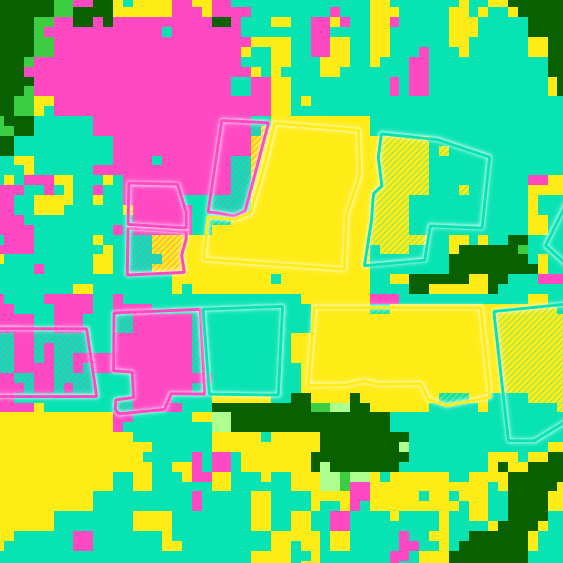} &
         \includegraphics[width=.2\columnwidth]{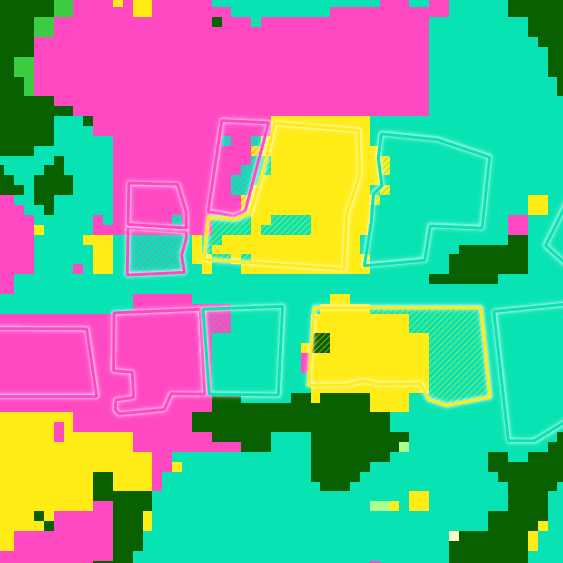} &
         \includegraphics[width=.2\columnwidth]{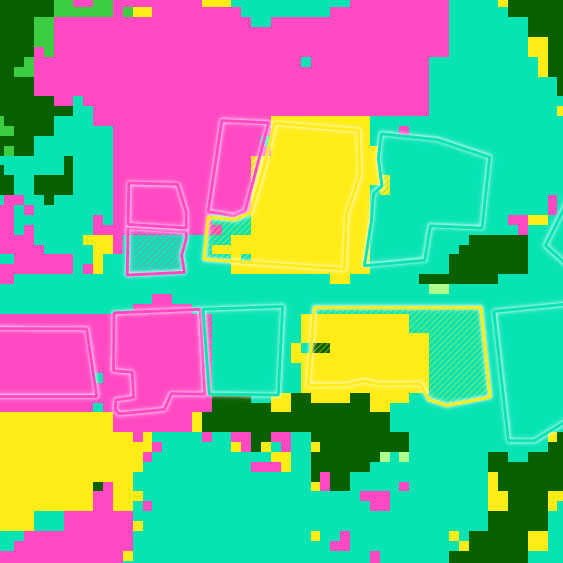} \\
    \end{tabular}
    \caption{Extracts from the provided land cover maps per method. Ground truth areas outlined over the extracts using the same color codes of Fig.~\ref{fig:koumbia_site}.}
    \label{fig:vis_insp}
\end{figure}

In order to give a further insight into the performances of the proposed method, we performed a qualitative analysis of the land cover maps provided by each of the competing methods. In Fig.~\ref{fig:vis_insp} we report some examples for the (2021 + 2020 $\rightarrow$ 2020) transfer scenario over the \firstsite{} study site. As easily observable, the \textit{Sourcerer} method tend to generate much noisier maps with respect to the competitors, which seems to be the main factor limiting its global performances. The other methods provide maps of comparable spatial characteristics, with \textit{REFeD} significantly outperforming the competitors on classes related to natural vegetation with different densities (like the \textit{Forest} class on the first row - bottom/right of the clip). More occasionally, \textit{REFeD} also seem to retrieve the correct crop class to whole fields within the cropland which are entirely misclassified by other methods (e.g. second row, for the big \textit{Cereal} field in the middle). Otherwise, the systematic improvement of \textit{REFeD} over its best competitors mainly occurs at the finest scales, like for the fields in the example on the last row of Fig.~\ref{fig:vis_insp}, where most of the ``holes'' generated by \textit{TempCNN} and \textit{POEM} appear as properly filled.

\subsubsection{Visualisation of internal model representations}
\label{ssec:features}

\begin{figure}
    \centering
    \begin{subfigure}{0.495\textwidth}
        \centering
        \includegraphics[width=\textwidth]{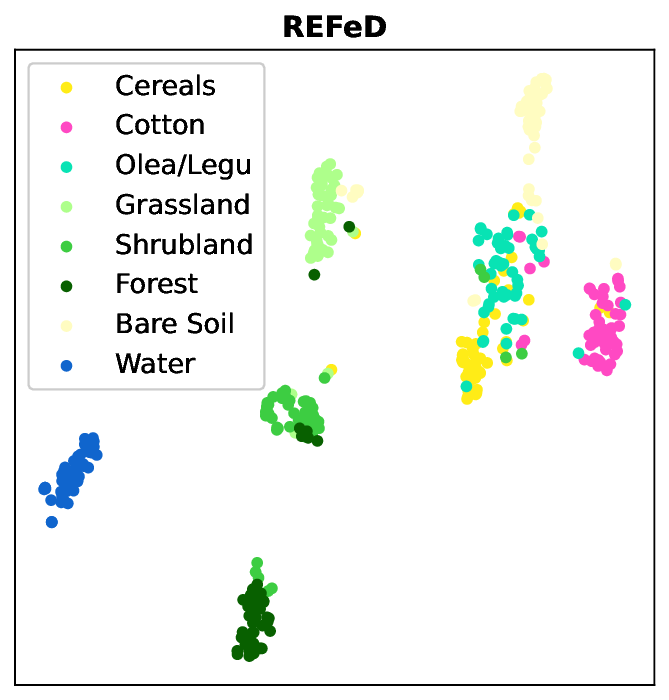}
    \end{subfigure}
    \hfill
    \begin{subfigure}{0.495\textwidth}  
        \centering 
        \includegraphics[width=\textwidth]{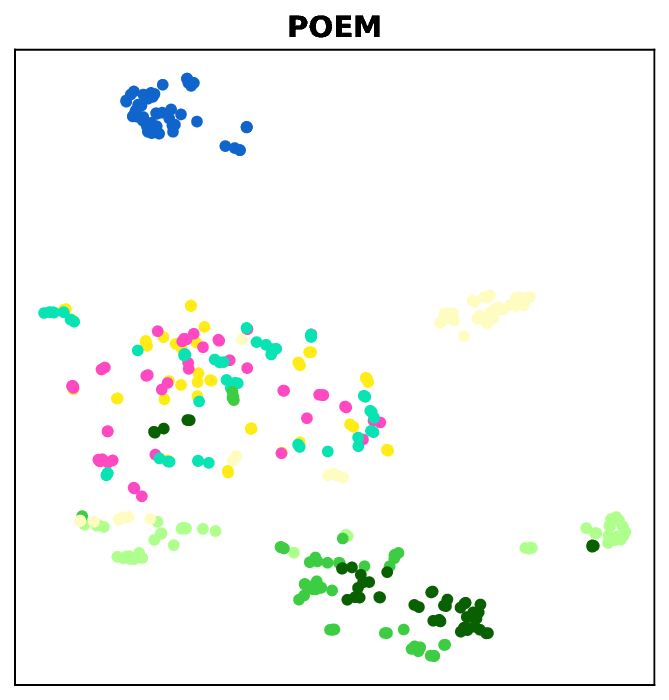}
    \end{subfigure}
    \begin{subfigure}{0.495\textwidth}   
        \centering 
        \includegraphics[width=\textwidth]{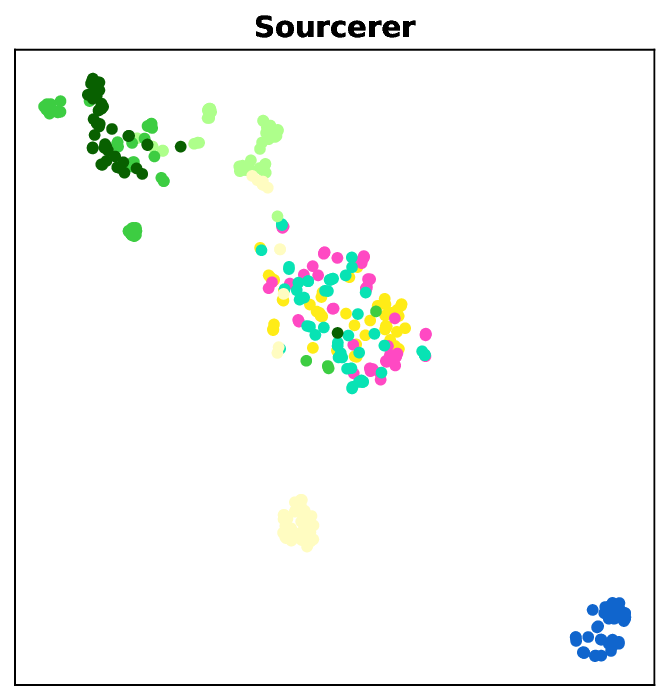}
    \end{subfigure}
    \hfill
    \begin{subfigure}{0.495\textwidth}   
        \centering 
        \includegraphics[width=\textwidth]{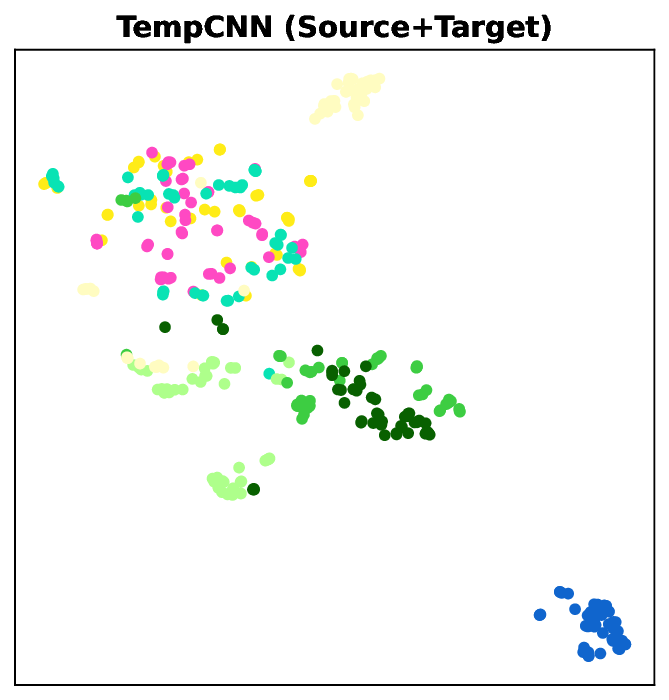}
    \end{subfigure}
    \caption{t-SNE results for the proposed approach (top left) and three different baselines: POEM (top right), Sourcerer (bottom left) and TempCNN (Source+Target) (bottom right).} 
    \label{fig:tSNE}
\end{figure}

In this last stage of our experimental evaluation, we provide a visual inspection of the internal feature representation learned by \method{}, \textit{POEM}, \textit{Sourcerer} and \textit{TempCNN} (Source + Target).
To this end, we randomly chose 50 samples per land cover class from the target domain and we extracted the corresponding feature representation per method. Subsequently, we applied t-SNE~\citep{Maaten2008VisualizingDU} to reduce the feature dimensionality for visualisation purposes. Results are depicted in 
Figure~\ref{fig:tSNE}.
We can note that all the methods well separate samples coming from the $Water$ and $Bare soil$ classes from the rest of the data. However, while competing approaches clearly mix samples from all other land cover classes together, \method{} partially alleviates clutter issues on the remaining classes providing a better visual behaviour in terms of cluster structure, on the considered subset of target data. This can be noted, for instance, regarding both the agricultural (\textit{Cereals}, \textit{Cotton} and \textit{Oleaginous/Leguminous}) and natural vegetation (\textit{Grassland}, \textit{Shrubland} and \textit{Forest}) classes. Overall, the visualisation of internal features representation is coherent with the quantitative as well as qualitative findings we previously discussed.

\section{Conclusion}
\label{sec:conclu}
In this work we have presented \method{}, a novel deep learning framework that enhances the accuracy of the current LC mapping process by combining together EO and reference data coming from two different domains (e.g. historical data and recent ones) with the aim to give value again to historical and/or overlooked reference data under a data-centric perspective. 

\method{} is based on a pseudo-siamese architecture with unshared weights and it relies on contrastive learning to disentangle invariant and specific per-domain features to recover the intrinsic information related to the downstream LULC mapping task. Furthermore, \method{} is equipped with an effective supervision scheme where feature disentanglement is further enforced via multiple levels of supervision.

The obtained results on two study areas covering extremely diverse and contrasted landscapes have highlighted the quality of our framework regarding both quantitative and qualitative analyses with respect to all considered competitors. Most importantly, \method{} systematically outperforms models that only exploit target data paving the way to the reuse of historical and/or overlooked reference data for the LULC mapping task taking as input satellite image time series data.

Potential future extensions related to our framework may include extending \method{} to encompass a multi-source remote sensing setting where the two domains are described by different sensors (e.g. Sentinel-2 and Landsat). Additionally, we could contemplate a scenario where multiple historical or out-of-year datasets are available further advancing the effective reuse of overlooked and/or neglected efforts related to past resource-intensive field campaigns conducted on the same or related study areas.


\noindent\textbf{Acknowledgements.} This research was funded by the French Space Study Center (CNES, TOSCA 2024 project) and the French National Research Agency under the Investments for the Future Program, referred as ANR-16-CONV-0004 (DigitAg). Dino Ienco acknowledges support for this work from the \textit{Agence Nationale de la Recherche} (ANR) under the ANR-23-IAS1-0002 (GEO-ReSeT) project.

\bibliography{refs}
\bibliographystyle{apalike}
\end{document}